\documentclass[conference]{IEEEtran}

\IEEEoverridecommandlockouts
\usepackage{cite}
\usepackage{amsmath,amssymb,amsfonts}
\usepackage{algorithmic}
\usepackage{graphicx}
\usepackage{textcomp}
\usepackage{url} 
\usepackage{xcolor}
\def\BibTeX{{\rm B\kern-.05em{\sc i\kern-.025em b}\kern-.08em
    T\kern-.1667em\lower.7ex\hbox{E}\kern-.125emX}}

\begin{document}

\title{BAUST Lipi: A BdSL Dataset with Deep Learning Based Bangla Sign Language Recognition\\
}
\author{\IEEEauthorblockN{1\textsuperscript{st} Md Hadiuzzaman}
\IEEEauthorblockN{2\textsuperscript{nd}Mohammed Sowket Ali}
\IEEEauthorblockN{4\textsuperscript{nd}Tamanna Sultana}
\IEEEauthorblockN{6\textsuperscript{nd}Abdur Raj Shafi}
\IEEEauthorblockA{\textit{Department of Computer Science and Engineering} \\
\textit{Bangladesh Army University of Science and Technology}\\
Saidpur, Bangladesh \\
hadiuzzaman908@gmail.com, sowket@gmail.com\\
tamannasultana189@gmail.com
}
\and
\IEEEauthorblockN{3\textsuperscript{rd} Abu Saleh Musa Miah}
\IEEEauthorblockN{5\textsuperscript{th} Jungpil Shin}
\IEEEauthorblockA{\textit{School of Computer Science and Engineering} \\
\textit{The University of Aizu}\\
Aizuwakamatsu, Fuukushima, \\965-8580, Japan \\
musa@baust.edu.bd\\
jpshin@u-aizu.ac.jp
}
}

\maketitle

\begin{abstract} People commonly communicate in English, Arabic, and Bengali spoken languages through various mediums. However, deaf and hard-of-hearing individuals primarily use body language and sign language to express their needs and achieve independence. Sign language research is burgeoning to enhance communication with the deaf community. While many researchers have made strides in recognizing sign languages such as French, British, Arabic, Turkish, and American, there has been limited research on Bangla sign language (BdSL) with less-than-satisfactory results. One significant barrier has been the lack of a comprehensive Bangla sign language dataset. In our work, we introduced a new BdSL dataset comprising alphabets totaling 18,000 images, with each image being 224x224 pixels in size. Our dataset encompasses 36 Bengali symbols, of which 30 are consonants and the remaining six are vowels. Despite our dataset contribution, many existing systems continue to grapple with achieving high-performance accuracy for BdSL. To address this, we devised a hybrid Convolutional Neural Network (CNN) model, integrating multiple convolutional layers, activation functions, dropout techniques, and LSTM layers. Upon evaluating our hybrid-CNN model with the newly created BdSL dataset, we achieved an accuracy rate of 97.92\%. We are confident that both our BdSL dataset and hybrid CNN model will be recognized as significant milestones in BdSL research. 

\end{abstract}


\section{Introduction}

People in the world use different languages to communicate with others, aiming to express their ideas, thoughts, requirements, basic needs, etc. There are various kinds of languages in the world based on their nature and social culture, like English, Hindi, Arabic, Turkey, German, Argentina, Japanese, Korean, Bangla, etc., which the common people use for their daily activities. Unfortunately, there are some hearing-impaired persons in our society either by birth or several affective diseases \cite{1islam2018ishara, 2tarafder2015disabling, 3begum2009computer}. According to statistics, there are 70 million hard-of-hearing and deaf people in the world, and nearly 3 million people in Bangladesh are currently using sign language \cite{4islalm2019recognition, 5fudickar2007user, 6karmokar2012bangladeshi,7bangladesh1994bengali, 8abedin2023bangla,9athitsos2008american,10ronchetti2016lsa64}. Hearing-impaired people cannot use spoken language; they need to use their body movements or signs to express their daily requirements and feelings. Sign language is a visual language or communication procedure based on human gestures, especially hand, facial, and body movements representing words to help people who cannot hear or talk \cite{11li2012sign,12alrecognition,13sandler2006sign,14islam2018potent,15murray2020intersectional}. It also can establish communication among the deaf and common people. Despite the sign language, the deaf and hard of hearing group has still been facing difficulties because of the lack of awareness, especially for education, medical services, employment, socializing and other daily equipment \cite{16miah2022bensignnet,17miah2023rotation,18miah2023dynamic,19miah2023multistage,20rahim2020hand}. However, sign language is a good communication medium for hearing-impaired people but is not easy to understand and learn. In addition, there is no universal or common sign language for the world, and usually, there is no mutual intelligibility among the sign language. Sometimes, there are different sign languages in the same language, like American and British sign language. To solve the above scenario problems, the deaf community needs an interpreter to establish communication with the common people for their daily activities. The problem is that getting an experienced translator is not easy because of the high cost and the high level of excellence. Automatic sign language recognition can play a crucial role in bridging the social and fundamental gap between the deaf and common people. In current decades, many domains have been used to develop sign languages, such as EEG based emotion for thinking class cation \cite{21miah2020motor,22miah2022natural,23zobaed2020real,24kibria2022bangladeshi,25joy2020multiclass,26miah2021event}, vision-based hand gesture recognition \cite{16miah2022bensignnet,27murray2020intersectional,28rafi2019image,29islam2018potent} and sensors \cite{31agarwal2013sign,32zhang2012kinect}. Like other sign language, most of the BdSL researchers have been working to develop a vision-based system.
\begin{figure*}[t] 
    \centering
    \includegraphics[scale=0.5]{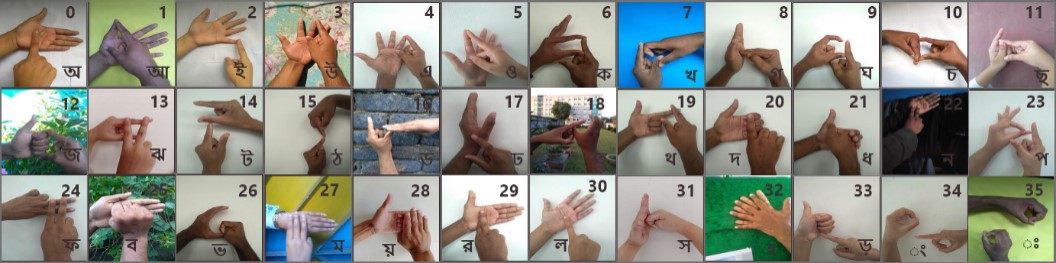}
    \caption{Dataset sample and labeling of each alphabet.}
    \label{fig1} 
\end{figure*}

Although the BdSL is one of the most important spoken languages in the world, especially in Bangladesh and India, very little research has been done on this language. Some researchers employed hidden Markov models (HMM), principal component analysis (PCA), and artificial neural network (ANN) to develop sign language recognizer \cite{33tarafder2015disabling,34shin2023korean,35oliveira2009skin}. Some researchers employed motion-capturing gloves in computer vision with multi-color gloves \cite{36zarit1999comparison,37sawant2014real}. More recently, some researchers applied deep learning-based convolutional neural networks (CNN) \cite{16miah2022bensignnet}. However, the progress of the BdSL is still not satisfactory as compared with other sign languages. The lack of a public dataset and the efficiency of the method are the main causes of the problem. To overcome the challenges, our contribution is to (i) create a new BdSL dataset and (ii) develop a new CNN to achieve satisfactory performance. We uploaded the code and data in the following link for use publicly https://shorturl.at/jklA9

\section{Literature Review}
\subsection{Different Sign Language Detection}
Although there are 300 sign languages worldwide, not all have a tool to recognize them. Computer vision-based researchers have been increasing interest in developing sign language recognition systems for different sign languages to make their lives easier. Most researchers have been working to develop American Sign Language (ASL). Among them, HMM is employed by many researchers to develop ASL \cite{37sawant2014real,38parton2006sign,39abedin2023bangla,40zafrulla2011american,41vogler1999parallel}. Sign language recognition also has been developed for Chinese sign language (CSL), Japanese sign language (JSL), French sign language (FSL), Arabic sign language (ASL), British Sign Language (BSL), Indian Sign Language (ISL), Turkey Sign Language (TSL) \cite{34shin2023korean,8abedin2023bangla,musa_norway}. ISL recognition system has been developed by \cite{42rajam2011real} to improve the ISL performance accuracy, Singha et al. applied Euclidean distance-based weighted eigenvalue \cite{43singha2013indian}. Hand tracking-based JSL recognition system has been developed by \cite{44imagawa1998color}, and a hand feature-based JSL recognition system has been proposed by \cite{45tanibata2002extraction}. A phoneme-based CSL recognition system has been submitted by \cite{46wang2002approach}, and a depth-based Bangla sign alphabet and digit dataset are proposed by\cite{rayeed2023bdsl47} some researchers employed CNN-based approach \cite{47vodopivec2016fine,48oyewole2018bridging} and other researchers are still working to develop a sign language recognition system for real-life applications.

\subsection{Bangla Sign Language (BdSL) Detection}
BdSL is quite unique compared to other sign languages and has not been explored among researchers. This language is inspired by ASL, ASL, and BSL, and many researchers used machine learning and deep learning approach to recognize BdSL\cite{rahim_poteto_leaves,egawa_fall,musa_multistream,electronics12092082,kabir2023investigating,9836398,Alzheimer_musa}. Hasan et al. created a new BdSL dataset by including 11 Bangla digits and word images \cite{49hasan2015new}. Then, they employed a universal interpreter for segmentation and feature extraction. K-nearest neighbor (KNN), Support vector machine (SVM), and artificial intelligence were previously the most commonly used algorithms for BdSL recognition \cite{50rahaman2014real}. Recently, researchers focused on the CNN to recognize BdSL instead of the machine learning-based algorithm. Shanta et al. employed a handcraft feature extraction technique with CNN to improve the performance accuracy of the BdSL recognition \cite{51shanta2018bangla}. Rafi et al. employed VGG19-based CNN to recognize BdSL \cite{28rafi2019image,52krizhevsky2012imagenet}. Islam et al. employed a deep learning-based CNN architecture to recognize BdSL digits \cite{13sandler2006sign}. A novel deep learning-based CNN model is proposed here, which has solved many existing problems and challenges in the BdSL domain.

\section{Dataset}
In this study, we introduced a new BdSL dataset comprising 18,000 frames, making it one of the largest BdSL datasets available. Our dataset encompasses a total of 18,000 images that represent 36 Bangla alphabets \cite{1islam2018ishara,3begum2009computer}. To compile this dataset, we gathered samples from 15 participants—both male and female. Ten of these participants were aged between 23-26 years, while the remaining five were between 40-50 years of age. It took 4 to 5 months of training for these subjects to produce the desired data. We utilized various smartphone cameras to capture these images, including but not limited to models like the Xiaomi Redmi 5, Mi Note 10, Samsung A10, and Nokia 6. Diversity was introduced in our dataset through various backgrounds: white, colorful, and natural. Moreover, we ensured a collection of 50 images for each alphabet captured under nighttime conditions.

\begin{figure*}[t] 
    \centering
    \includegraphics[scale=0.4]{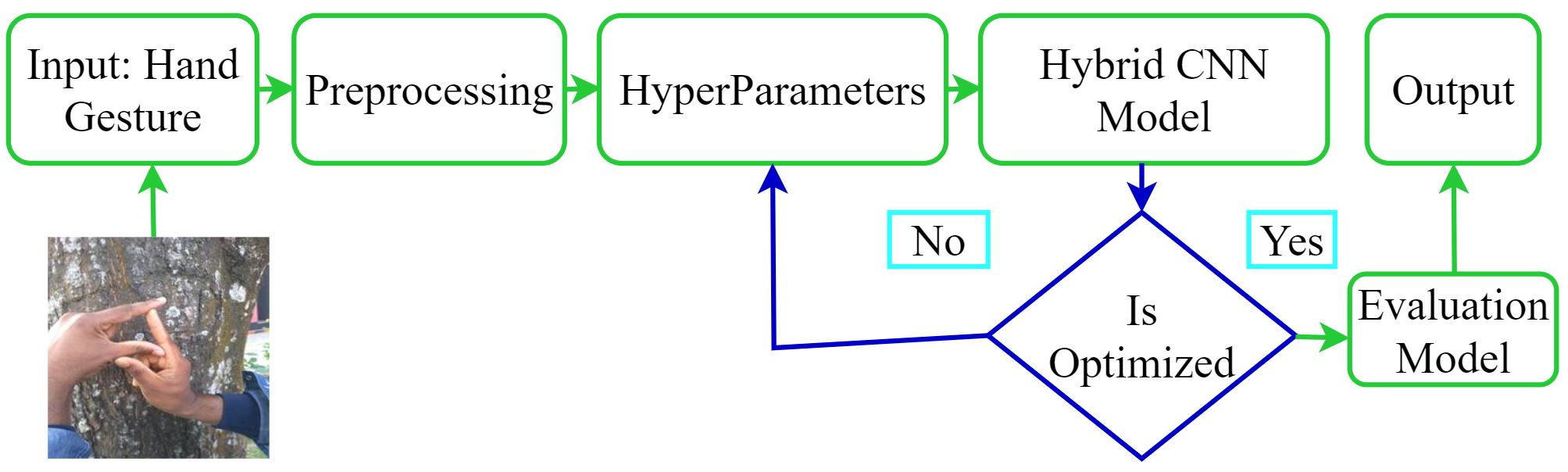}
    \caption{The working flow diagram of the proposed BdSL recognition system.}
    \label{two} 
\end{figure*}

Before embarking on data collection, we ensured to obtain the consent of all participating subjects and also received the necessary approval from the Bangladesh Army University of Science and Technology (BAUST). To simplify the utilization of our dataset in Machine Learning, Deep Learning, and Computer Vision domains, we meticulously organized the data of the 36 alphabets into individual folders. These folders follow a sequential naming pattern, labeled from 0 through 35. Each folder designation corresponds directly to a specific Bangla alphabet, with every folder housing 600 images that symbolize its respective alphabet. Figure \ref{fig1} provides a glimpse into the samples from our proposed BdSL dataset, while Figure \ref{two} delineates the Bangla Sign alphabet alongside the labels for each. For broader accessibility and ease of use, all the accumulated images have been stored on an HDD hard disk and subsequently uploaded to GitHub for public access.

\section{Proposed Model Construction}
The working flow architecture and major steps of the study have been summarized in Figure \ref{two}. Our research primarily revolves around the development of a robust recognition system for Bangla sign language. The methodology we've adopted for this task can be segmented into three distinct phases: Preprocessing, Hybrid Deep Learning Model, and Hyperparameter Optimization. In the preliminary phase, raw images from the dataset undergo a rigorous preprocessing routine. This routine ensures normalization of image sizes, histogram equalization for uniform brightness, and noise reduction, among other techniques. The aim is to ensure that the images are in an optimal state for feeding into the deep-learning model. In the second phase, the images are introduced to our hybrid model. This model is an innovative fusion of deep learning layers, which captures intricate patterns and features from the image, coupled with an LSTM (Long Short-Term Memory) model, adept at recognizing and remembering sequences over extended periods. Such a structure makes our model particularly suited for sign language recognition, given the sequential and patterned nature of sign gestures.     In the final phase, we harness several hyperparameters to fine-tune and optimize our model's performance. Techniques like grid search and random search are deployed to determine the ideal values for parameters like learning rate, batch size, and dropout rate, ensuring that the model is both effective and efficient. Through this methodology, we aim to attain a Bangla sign language recognition system that is both accurate and swift, contributing a valuable tool to the field.

\begin{figure*}[t] 
    \centering
    \includegraphics[scale=0.28]{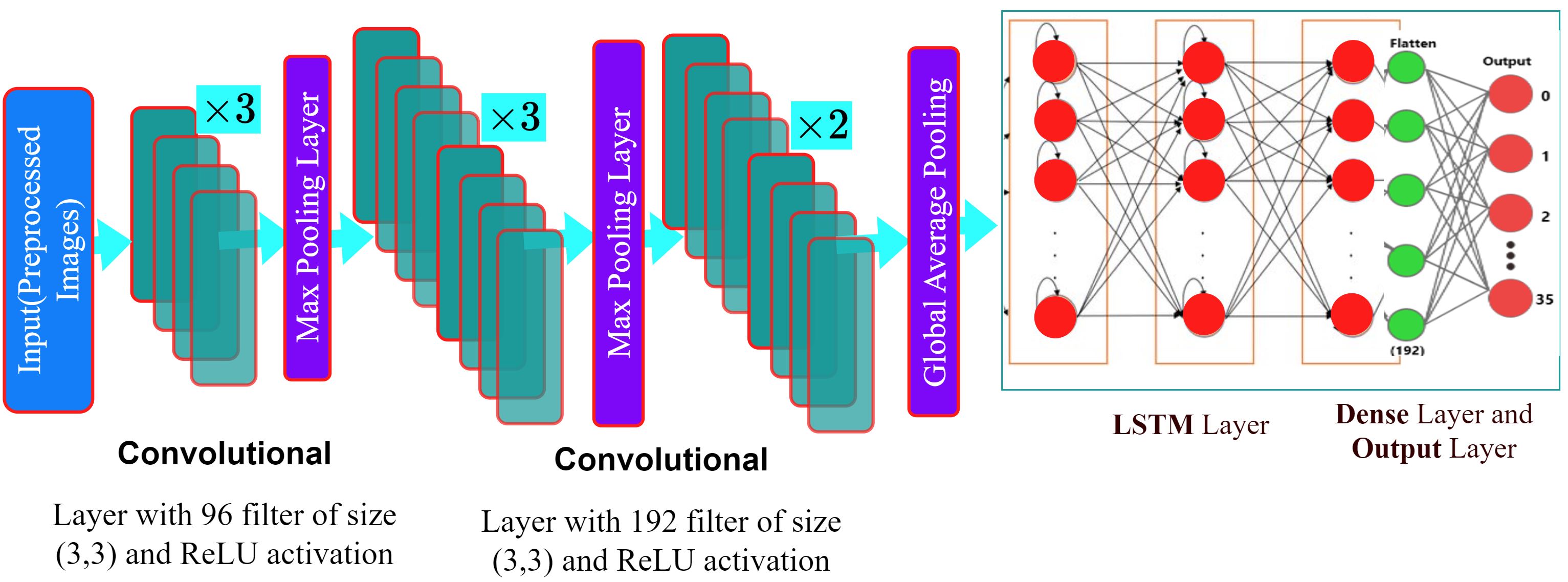}
    \caption{Proposed Hybrid-Model Architecture.}
    \label{fig2} 
\end{figure*}

\subsection{Data Preprocessing}
Before the CNN, we normalized the RGB image after converting it into a gray scale by dividing it with the maximum pixel and resizing the image to 50 x50.
\subsection{Hybrid Deep Learning Model }
In the second phase, the images are introduced to our hybrid model. This model is an innovative fusion of deep learning layers, which captures intricate patterns and features from the image, coupled with a Long Short-Term Memory (LSTM) model, adept at recognizing and remembering sequences over extended periods. Such a structure makes our model particularly suited for sign language recognition, given the sequential and patterned nature of sign gestures. In our research, we meticulously crafted a system to handle data characterized by a matrix-like structure, such as imagery. Our model is structured in two main parts: initially, we utilize a Convolutional Neural Network (CNN), and subsequently, we employ a Long Short-Term Memory (LSTM) network. CNNs are essentially a cascade of numerous layers of interlinked nodes. These encompass an introductory input layer, several intermediate hidden layers, and a concluding output layer. These intermediate, or hidden, layers are segmented into multiple convolutional and pooling layers. Every convolutional layer is designed to deploy a unique set of filters onto the incoming data. Each individual filter is specifically tailored to pinpoint a particular attribute or feature in the data. The subsequent pooling layers are integral for dimensionality reduction. They perform this by compressing the convolutional layers' output, often through a down-sampling technique. Once the CNN has processed the data, it then navigates this data through a densely linked layer. This fully connected layer assumes the pivotal role of categorizing the processed input data, as illustrated in references \cite{3begum2009computer,5fudickar2007user,35oliveira2009skin}. 

In the succeeding section, we aim to delve deeper, exploring the intricate operations of each layer as they contribute to the task of sign classification. Our architecture boasts eight convolutional layers, paired with two pooling layers, three dropout layers for regularization, two dense or fully connected layers, a couple of activation layers, and a final output layer to present the results. To further illustrate, our premier convolutional layer is characterized by 92 distinct filters, each with a kernel dimension of (3x3). Additionally, it employs 'same' padding and activates using the ReLU(Rectified Linear Unit) function. Then reduced the overfitting of data used by 20\% dropout. The 2nd and 3rd convolution layers also have 92 filters, kernel size (3x3), activation ReLu, and padding. But in 3rd convolution layer, stride 2. To reduce overfitting, we once again applied a 50\% dropout. A (2x2) pooling layer was added next. In the 4th, 5th, and 6th convolution layers, we used 192 filters, keeping the other parameters the same as before. An additional dropout layer was introduced to further prevent overfitting, followed by another (2x2) pooling layer for the 7th convolution layer. The 8th convolution layer has 192 filters, kernel size (1x1), and padding valid. Then, there is batch normalization, global max pooling, a fully connected layer, and finally, the output layer with filter 36 with activation function softmax. 
Input Layer. Using the input layer, we fed the preprocessed data into the network, and there are 2500 nodes contained by the input layer for $50 \times 50$ input images. Convolutional Layer. The output of the input layer is fed into the convolutional layer where it performs a convolution between n x n x d input dimension where n is height, width, d defined depth of the image and M kernels with k x k pixel size. The filter traveled from left to right of the input image by performing multiplication pixel by pixel. Finally, to protect from the shrinkage problem, we added zero padding around the image pixel \cite{3begum2009computer,5fudickar2007user}. \\
\\
\textbf{Pooling Layer}\\
In the architecture, we strategically incorporated two distinct types of pooling layers. The first one is the max-pooling layer, and the second is the global average pooling layer. By applying these layers, we were able to efficiently dissect the n x n feature map into several distinct n segments. For each individual segment, the max-pooling layer was designed to meticulously select the peak value, leading to the generation of a feature that is roughly half the size of its original counterpart. On the other hand, for the purpose of our study and to achieve a more streamlined representation, we opted to employ a global average pooling layer. This layer plays a pivotal role in transforming the detailed feature matrix into a more manageable vector format, as cited in references \cite{3begum2009computer,5fudickar2007user}.\\
\\
\textbf{Overfitting and Underfitting Control Layers.} \\
We employed the architecture batch normalization and dropout layer to protect the architecture from overfitting and underfitting, which also worked here as a regularisation. During the training phase, by measuring the probability value, the dropout layer randomly identifies the ignoring of the neuron. During the forward pass or backward pass, ignoring neurons does not allow them to move anywhere. On the other hand, the batch normalization layer reduced the training of the model by calculating the minibatch in each trial \cite{3begum2009computer,5fudickar2007user}.\\
\\
\textbf{Activation and Loss Function}\\
To normalize the output of the convolutional layer, we employed the activation function, and ReLU is used here, which works as a non-linear function. On the other hand, the loss function is used as a loss function categorical cross-entropy (CCE)\cite{53de2005tutorial} for producing bad and good results. In explanation, it will produce a bad result, in which it got 0.011 prediction probability against one actual probability. The perfect cross entropy value and minimize score of this function is 0 and considered a bad result when a probability is 0.0 \cite{3begum2009computer,5fudickar2007user}. \\
\\
\textbf{LSTM Layer}\\
Long Short-Term Memory (LSTM) is a type of recurrent neural network (RNN) optimized for sequence-based tasks, making it suitable for sign language recognition, which involves sequential video frames or motion data. Each frame in a video sequence becomes an input to the LSTM, allowing it to track the evolution of gestures over time. LSTMs achieve this through three main gates:
\begin{enumerate}
\item \textbf{Forget Gate:} Decides which information to discard from the cell state.
\item \textbf{Input Gate:} Updates the cell state with new information.
\item\textbf{Output Gate:} Determines the next hidden state based on the current input and cell state.
For sign language recognition, this memory ability is vital, as context from previous frames can influence the interpretation of subsequent frames. In practice, the LSTM layer might be preceded by convolutional layers for feature extraction or an embedding layer for structured data. For complex patterns, multiple LSTM layers can be stacked, improving recognition accuracy. After the LSTM processes the data, an output layer predicts the specific sign or gesture.
A challenge with this approach is the variable duration of sign gestures, requiring mechanisms like padding to handle different sequence lengths. For real-time applications, model complexity must also be balanced with performance speed.
In sum, LSTMs, with their sequential data processing capability, offer a promising avenue for effective sign language recognition when combined with other neural network components.
\end{enumerate}

\textbf{Fully Connected Layer}\\
 Finally, we employed a GAP layer for the final feature, which is fed into the fully connected (FC) layer. The main concept of the FC layer is that it is considered the convolutional layer, a 1*1 filter size. Then make a link between every neuron in one layer with the other layer neurons. Here it is performed by the vector extracted by the previous gap layer \cite{3begum2009computer,5fudickar2007user}.  
 
 \textbf{Output Layer}
 We considered the final layer as an output layer that compared the output with the corresponding class number. To perform this, we employed a SoftMax function. 

\section{Performance Evaluation}  
\subsection{Environment Setting}
All images in the BAUST LIPI dataset are $224 x 224$ pixels and uploaded to Kaggle. To reduce computation time, we resize it to 50 x 50 pixels for use in machine learning and deep learning models and divide our dataset into three sets: training, validation, and testing. Among 18000 images, we used 12960 images for training, 3240 images for validation, and 1800 images for testing, where 72\% training, 18\% validation, and 10\% testing, respectively. There are four algorithms that we apply to our dataset. The construction model is shown in Figure \ref{two}. In our CNN model, we use 0.0001 learning rates with Adam optimizer.
\subsection{Performance Evaluation}
Table \ref{table1} highlights the superior performance of the proposed CNN-SLTM model compared to other algorithms. While the Decision Tree and SVM models achieved testing accuracies of 62.08\% and 73.85\%, respectively, and the CNN model reached 80.44\%, the CNN-SLTM model excelled with a remarkable testing accuracy of 97.28\%. This improvement is due to the hybrid architecture of CNN-SLTM, which combines CNN's powerful feature extraction capabilities with SLTM's ability to capture long-term dependencies, leading to more accurate predictions and robust generalization across training, validation, and testing datasets. The model's 99.91\% training and 97.29\% validation accuracies further demonstrate its effective machine learning algorithm, including the decision tree and svm. 
 \begin{table}[h]
	\caption{Performance evaluation with proposed model}
	\label{table1}
	\centering
	\begin{tabular}{llll}
		\hline
		Algorithm                     & Training      & Validation        & Testing \\ \hline
  		 Decision Tree             & -  &- & 62.08\%\\
    \hline
		  SVM             & -  &- & 73.85\%\\
    \hline

    		  CNN             & 99.71\%  & 81.00\% & 80.44\%\\
    \hline
    		Proposed  CNN-SLTM             & 99.91\%  & 97.29\% & 97.28\%\\
    \hline
		
	\end{tabular}
\end{table}

\begin{figure}[h] 
    \centering
    \includegraphics[scale=0.50]{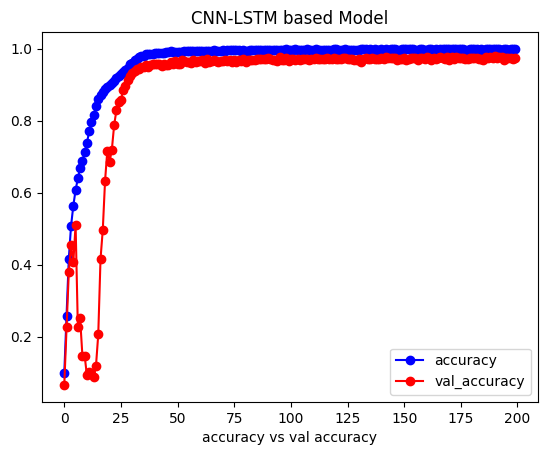}
    \caption{Training and Validation Accuracy Curve}
    \label{fig4} 
\end{figure}
\begin{figure}[h] 
    \centering
    \includegraphics[scale=0.50]{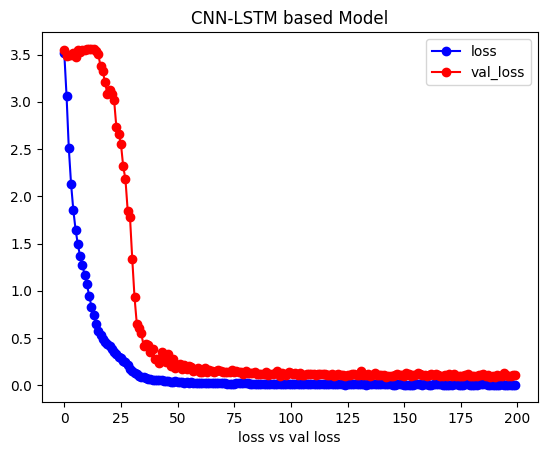}
    \caption{Training and Validation Loss Curve}
    \label{fig5} 
\end{figure}

\section{Conclusion and Future Works}
In this study, we developed a BdSL dataset, namely BAUST LIPI, and validated it with a new proposed CNN-LSTM hybrid model. Our work is also practical for recognizing different signs of the BdSL precisely, and it will solve the trouble for the deaf and mute community who cannot skill with BdSL to establish communication among themselves without taking any interpreter. The generalization capacity in any model can serve in a broader research domain, such as in the case of BdSL automatic recognition. In work, we mainly develop a new BdSL dataset to overcome the dataset limitation problem in the BdSL research domain, and it may decrease the limitation in upcoming research. In the future, we are planning to utilize the proposed method with other existing datasets for BdSL and other languages, aiming to build a real-time BdSL recognition system. 

\section{Acknowledgements }
This work was supported by the Computer Science and Engineering Department of Bangladesh Army University of Science and Technology (BAUST) for collecting datasets. Those people helped us build the dataset: Md Asaduzzaman Sarker, Fahima Akther, Binoyee, Tuli, Sonchita, Afifa Tasnim, Fariha Sultana, Rezoana Bonna, Najmun Oishi, Mantasha Mustarin, Mou, Shaiduzzaman, Arif, etc.
\bibliography{mybib}
\bibliographystyle{unsrt}

\vspace{12pt}

\end{document}